\documentstyle[epsf,General9705,dir,bib_cite,Local,%
camera_ready%
]{article}
\def\EJ#1#2{#1}
\def\tcomment#1{#1}
\def\comment#1{#1}
\def\Scomment#1{}     

\begin{document}
\setcounter{page}{95}
\singlespace\rnormalsize

\Fdouble{\doublespace\rnormalsize}
\baselineskip=4.53mm

\def\tcomment#1{}
\def\comment#1{#1}

\Fdouble{\singlespace\rnormalsize}



\title{%
Topic Graph Generation for Query Navigation:\\
Use of Frequency Classes for Topic Extraction}

\author{Yoshiki Niwa, Shingo Nishioka, Makoto Iwayama, \and Akihiko Takano\\
Advanced Research Laboratory, Hitachi, Ltd.\\
Hatoyama, Saitama {\tt 350-03} Japan\\
{\it \{yniwa, nis, iwayama, takano\}@harl.hitachi.co.jp}
\AND
Yoshihiko Nitta\\
Dept. of Economics, Nihon University\\
1-3-2 Misaki-ch\^o, Chiyoda-ku, Tokyo {\tt 101} Japan\\
{\it nitta@eco.nihon-u.ac.jp}
}

\maketitle

\Fdouble{\doublespace\rnormalsize}

\begin{abstract}\it
%
%

\EJ{%
To make an interactive guidance mechanism for document 
retrieval systems, we developed a user-interface which presents users 
the visualized map of topics at each stage of retrieval process.
Topic words are automatically extracted by frequency analysis and 
the strength of the relationships between topic words is measured 
by their co-occurrence.
}{%
-
-
-
-
-
}%
\EJ{%
A major factor affecting a user's impression of a given topic word
graph is the balance between common topic words and specific topic words.
}{%
-
-
}%
\EJ{%
By using frequency classes for topic word extraction,
we made it possible to select well-balanced set of topic words, 
and to adjust the balance of common and specific topic words.
}{%
-
-
}

\Comment{
A prototype system was built for CD-ROM version of Nikkei Newspaper'94 
which contains about 180,000 articles.
Experimental use of this system supports the usefulness of this
guidance by topic-word maps.
}

\Comment{
}

\end{abstract}

\tcomment{\setcounter{section}{0}\def\tcomment#1{}}

\section{Introduction}

\tcomment{\setcounter{section}{0}\setcounter{page}{1}
\def\tcomment#1{}
\section{\EJ{%
Introduction
}{%
-
}}
}

\EJ{}{
\subsection{-}
}

\EJ{
As the rapid development of networks brings us easy access
to large quantities of on-line information, the need to find ways of
retrieving useful information is increasing.
Retrieval is not always an easy task, however, 
because an inquiry that is imagined by a person cannot 
be directly conveyed to a machine, 
and because inquiries themselves are not always clear.
A guidance function which supports an interactive approach to 
the target information is therefore required, and
wwe have developed a guidance system that presents to users 
a visualized map of topics at each stage of interaction.
}{
-
-
-
-
-
-
-
}

\EJ{
}{
-
-
-
-
-
(\cite[p.1263-4]{Fujisawa+Kinukawa:IPSJ93})-
(\cite[(-) p.111-114]{Ingwersen:93})
}

\EJ{
This is done by automatically constructing graphs
of topic words in the retrieved documents.
Topic words are extracted by frequency analysis
and the strength of the relationships between topic words is 
measured by their co-occurrence.
}{
-
-
-
-( \cite{Niwa:KK96-606})
}

\Figsw{
\vskip3\rmm
\centerline{\figureformEPSFab{75}{63}{fig_IRIR_GE.epsf}{\fignumIRIR}{Interactive IR}
}
}{\centerline{(Fig. \fignumIRIR)}}

\EJ{
Figure \fignumIRIR\ illustrates the flow of our interactive retrieval system.
After a user starts with a query,
a set of documents is retrieved and displayed as a list of their titles.
From these retrieved documents are extracted topic words 
along with their mutual relationships,
and these words and relationships are displayed in a graph structure.
From the title list we can get concrete information about the
retrieved results; for example, ``documents concering such and such
topics were retrieved.''
From the topic graph, on the other hand, we get more abstract
information like ``such and such topics are dominant in the retrieved
documents.''
By referring to both concrete and abstract information concerning the
results, users can proceed to the next trial with a better perspective.
}{
{-\fignumIRIR}
-
-
-
-Halley  Hyakutake 
-
Shoemaker Levi -
-
-
}


\tcomment{\setcounter{section}{1}\def\tcomment#1{}}

\section{Related Work}

\tcomment{\setcounter{section}{1}\setcounter{subsection}{0}
\def\tcomment#1{}
\section{Related Work}
}

\Scomment{
Interactive user-interface for information retrieval has
two complementary factors, guidance and feedback.
This research is primarily concerned with the guidance factor.
Relevance feedback introduced by Salton \shortcite{Salton:1968}
is wellknown as the latter.}

Visualization of information has been attracting 
a lot of interest, and many studies have been done 
(Rao et al. \shortcite{Rao+al:CACM95}).
Some of them visualize the document space and others visualize
the lexical space. 
Our study does the latter.
A closely related work is the scatter-gather method developed by
Cutting {\em et al.} \shortcite{Cutting+al:SIGIR92}.
In their method, retrieved documents are automatically clustered 
according to the similarity of their document vectors, 
then the characteristic words of each cluster are extracted.
One problem with this method is the computational complexity of 
clustering, which is about the third power of the number of documents.
\Comment{
-
-
}
In the development of our system, we paid attention to the 
real-time response and therefore to the computational tractability.

Another active issue in lexical visualization is the merger of a
thesaurus database and automatically extracted topic words.
This issue has been studied by the Illinois University group 
in their digital library project
\cite{Johnson+Cochrane:DL95,Schatz+al:DL96}.
The creation of additional effects beyond the sum of two factors
may be a future problem.

Visualization of document spaces seems to be more popular than that of
lexical space.
In the Smart system \cite{Salton+al:Science94},
text relation maps are constructed according to the similarity 
of their vector representation.
Xerox's Information Visualizer system \cite{Mackenlay:CHI95},
visualizes the reference structure of documents.
One major problem with the visualization of document space is
the representation of documents.
To give proper information to users, document titles need to be presented.
However, they are not always compact.
This is where the visualization of lexical space is advantageous.

In Japan, visualization of lexical space has primarily been studied in the
framework of idea processing systems by the group at Tokyo University
\cite{Sumi+al:IEICE95E}.
Recently, Sugimoto \shortcite{Sugimoto+al:SIGNL96E} proposed a method
of displaying documents and related words at the same time.
Interaction of documents and related words should be a goal for the 
future research.
There are also some research studies concerning the document browsing 
system using the visualization technique
\cite{Morohashi+al:ISDL95,Arita+al:SIGNL95E}.


\tcomment{\setcounter{section}{2}\setcounter{page}{3}
\def\EJ#1#2{#1}
\def\tcomment#1{}}

\section{\EJ{%
Generation of Topic-Word Graphs
}{%
-
}}

\EJ{%
Our topic graph generation method consists of following three steps.
}{%
-
}

\begin{enumerate}
\EJ{%
\item Topic word extraction.
\item Link generation by co-occurrence analysis for extracted topic words.
\item Graph mapping to a 2-dimensional area.
}{%
\item -
\item -
\item -
}
\end{enumerate}

\EJ{
An outline of the method has already been described \cite{Niwa+al:ICCPOL97}.
Here we explain the details with reference to Fig. \fignumTWGmethod .
}{%
-\cite{Niwa:KK96-606}
-
-
-\fignumTWGmethod 
}

\Figsw{
}{\centerline{(Fig. \fignumTWGmethod)}}

\subsection{\EJ{%
Topic word extraction
}{%
-
}}

\tcomment{\setcounter{section}{2}\setcounter{subsection}{0}
\def\EJ#1#2{#1}
\def\tcomment#1{}
\subsection{\EJ{%
Topic word extraction
}{%
-
}}
}

\EJ{%
Topic word extraction is based on the importance of each word
appearing in the retrieved set of documents.
We measure the importance of each word by the ratio of $df$ to $DF$,
where $df$ is the number of retrieved documents containing the word
and $DF$ is the number of all documents containing the word in the
entire target database.
We call this ratio the relative frequency of the word.
This importance measurement is an embodiment of a natural idea that
if a word not so common in general appears frequently in the
retrieved results,
then the word is probably characteristic to the retrieved set of
documents.
}{%
-
-
-
-
-
-
}

\EJ{%
Figure \fignumTWGmethod\  shows the case where the query is
\q{global environment}.
The word \q{greenhouse}, for example, appears in 62 documents in the
retrieval results and in 268 documents in the whole database.
So the relative frequency is 62/268 = 0.23.
We calculate the importance (relative frequency) of all words
appearing at least once in the retrieved documents and select
some of the most important words.
}{%
-\fignumTWGmethod\ ''commet''Shoemaker  
- (df=24), 
- (DF=24)  df/DF=1 
- Halley 
df/DF=15/19=0.79 -
-
}

Relative frequency is an intuitively acceptable measurement of
importance and has the additional merit of not being affected
very much by random sampling of retrieved documents.
It has a problem, however, in that it suffers from noises caused by
low frequency words. 
The improvement of this simple method is given in section 4.

\Scomment{
\EJ{%
For example, if a rare word appearing only once in the entire database 
accidentally appears in the retrieved results, 
then the relative frequency takes its maximum value 1.
And such cases are not exceptional.
To avoid such noises, we adopted a method of classifying words into
frequency classes and taking topic words from each of the 
frequency classes.
The details of this method are given in section 5.
}{%
-
-
-
-
-
-
-
-
-
-
-
-}
}

\Comment{
-
-
}

\Comment{
(-
-
-
-
}

\subsection{\EJ{%
Link generation by co-occurrence analysis
}{%
-
}}

\tcomment{\setcounter{section}{2}\setcounter{subsection}{1}
\def\EJ#1#2{#2}
\def\tcomment#1{}
\subsection{\EJ{%
Link generation by co-occurrence analysis
}{%
-
}}
}


\def\tmpa{\q{ozone}}\def\tmpb{\q{dioxide}}

\Scomment{
\EJ{%
The second step is to generate links between topic words
by using co-occurrence statistics to measure their 
inter-relationship.
The table in the upper part of Fig. \fignumTWGmethod\ is the
co-occurrence table of topic words.
Each position of the tabel has two numbers.
The upper number is the co-occurrence frequency $f_{xy}$ of two 
words X(column) and Y(row);
that is, the number of documents in the retrieved results which
contain both X and Y. 
The lower number shows the co-occurrence strength of Y with respect to X. 
We measure this strength as the co-occurrence frequency $f_{xy}$ 
divided by the frequency of Y.
In case of {\tmpa} and {\tmpb}, for example, their 
co-occurrence frequency is 19 and the co-occurrence strength of
{\tmpb} with respect to {\tmpa} is 19/48 = 0.40.
}{%
-
-{\fignumTWGmethod}
-
-
-
-
-
-
-''shoemaker''  ``solar'' 
7 - shoemaker  solar  7/30=0.06
-
}
}

\EJ{
\Scomment{
The next step is the generation of links for drawing topic word graphs.
}
In our method, each topic word X is linked to a word
which has the highest co-occurrence strength with respect to X
among the topic words having higher frequency than X.
The co-occurrence strength of Y with respect to X is measured
by the number of documents containing both X and Y by the
number of documents containing Y.
In case of \tmpa, the words having higher frequency are
\q{global}, \q{greenhouse}, and {\tmpb}.
Of these three words, {\tmpb} has the highest co-occurrence
strength with respect to {\tmpa}, 0.40.
Therefore, {\tmpa} is linked to {\tmpb}.
By applying this process to all topic words, we get the link table
(at the right of the co-occurrence table).
}{%
-
-
-
-''Shoemaker'' 
comet, solar, asteroid -
comet - 0.23 ''shoemaker''  comet 
-
-
}

\Figsw{
{\Columnsw{\rmm=1.7mm\unitlength=1.0\rmm}{}
\vskip5mm
\hspace{-2mm}

{\Columnsw{\rmm=1.7mm\unitlength=1.0\rmm}{}
{\rmm=1.4mm\unitlength=1.0\rmm
\figureformEPSFab{75}{52}{fig_TWGmethod_GE.epsf}{\fignumTWGmethod}{%
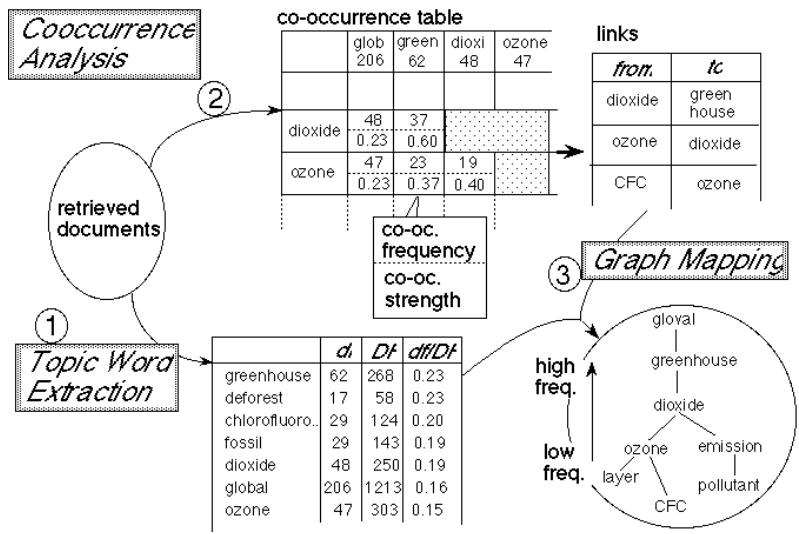
\EJ{TWG Generation Method}{
-
}}

}
}

\vspace{-5mm}
}
}{}

\subsection{\EJ{%
Graph mapping
}{%
-
}}

\tcomment{\setcounter{section}{2}\setcounter{subsection}{2}
\def\tcomment#1{}
\subsection{\EJ{%
Graph mapping
}{%
-
}}
}

\EJ{
In the first step we got a set of topic words which form the
nodes of a topic word graph to be generated, 
and in the second step we got links between these topic words.
Thus we now have the topological structure (nodes and links)
of the topic word graph. 
In this step we map the graph to a 2-dimensional rectangular 
area of given size.
}{
-
-
-
}

\EJ{
The mapping is done by determining the xy-coordinates of each node.
Again, there are many possible methods.
In what follows we describe the method currently used in our prototype 
system.
}{
-
-
-
}

\EJ{
The y-coordinate of a node is first derived from the document frequency
($df$) of the corresponding topic word by using the following formula:
\(y\;=\;C_1\;\tan^{-1}\left(C_2\;log\,(df\,/\,df_m)\,\right).\),
where, $df_m$ is the middle frequency of all of the topic words.
}{
-(df)
-
-
-
-
-
-
-
-
-
-
-
}

The terms $C_1$ and $C_2$ are constants.
Then x-coordinates are allocated recursively starting from the
top frequency node, just to prevent overlaps.

\Comment{
\EJ{
Secondly, x-coordinates were allocated recursively starting with
nodes having no parents. Their x-coordinates are the points equally
dividing the given region's horizontal interval.
In the recursive process, we take a set of nodes sharing a same
set of parent nodes for which x-coordinates are already determined.
Let [a b] be the horizontal interval of given region and p be
the average of the parents' x-coordinates. In case p is larger than
the center, (a + b)/2,  the x-coordinates of the nodes are the 
points equally dividing the interval [2p-b b].
}{
}}


\Scomment{\input{4}}  

\tcomment{\setcounter{section}{4}\def\EJ#1#2{#2}\def\fcomment#1{}
\def\EJ#1#2{#1}
\def\tcomment#1{}}

\section{
\EJ{%
Topic-word extraction using frequency classes
}{%
-
}
}

\EJ{
The relative frequency of a word is a natural index for measuring how
prominent a word is in a particular set of documents.
If topic words are selected according only to this relative
frequency, however, we usually miss some important high-frequency words.
}{
-
-
-
}

\leftskip=28mm
\EJ{
When the topic words of a retrieved document set are displayed for
characterizing the document set,
the balance of common and specific topic words is important.
This balancing problem cannot be solved by using other criterion
for measuring the importance. For example, the {\em tf*idf} measurement
is advantageous for selecting high frequency words but misses some 
important specific topic words.
We therefore used a topic word extraction method based on the 
classification of words by their document frequency (number
of documents containing a word) in the retrieval results.
}{
-
-
-
-
-
-
-
}

\Comment{
\EJ{{\bf Effect:}}{{\dg -}}
\EJ{
By this method, we can take high frequency topic words and low frequency.
and we can tune the balance of these different frequency classes.
Merits of using frequency classes:
\begin{itemize}
\item We can always take well balanced set of topic-words 
\item We can change the weight of 
\end{itemize}
}{%
-
-
}
}

\Comment{
\subsection{\EJ{%
Balance of common topics and specific topics
}{%
-
}}
\input{5-1}
}

\Comment{
\subsection{\EJ{%
Details on topic-word extraction
}{%
-
}}
}

\tcomment{\setcounter{section}{5}\setcounter{subsection}{1}
\def\EJ#1#2{#1}\def\fcomment#1{}
\def\tcomment#1{}
}

\leftskip=0mm
\EJ{
In the framework of our topic word extraction method using
frequency classes, we use the following parameters for
setting the frequency thresholds defining the frequency classes
and the allotment of number of topic words to each frequency class.
}{%
-
-
-
}

\begin{itemize}
\EJ{%
\item[$N$] Total number of topic words to be extracted
\item[$C$] Number of frequency classes
\item[$L$] The lower frequency boundary
\item[\paramB] Balance parameter
}{%
\item - (N)
\item - (C)
\item - (L)
\item - (\paramB)
}
\end{itemize}

\subsection{\EJ{%
Thresholds of frequency classes
}{%
-
}}

\EJ{%
The first stage is the definition of frequency classes,
and the parameters $C$ and $L$ are used for setting the frequency
thresholds.
The lower frequency boundary $L$ takes a value within the interval
(0 1], and words less frequent than $L\times M$ are excluded from
the candidates of topic words. Here $M$ is the maximum document
frequency of a word in the retrieved documents over the set of all
words appearing in the retrieved documents.
}{%
-
-
-
-($L$) (1/32) $\times\ L$
-
}

\EJ{%
The frequency classes are determined by parameters $C$ and $L$ in the
following way. Here $df$ means the document frequency of a word.
}{%
- ($C$)  ($L$) 
-
- $M$ 
}

\[\hbox{\EJ{Class}{-}}\ k: M\ r^{k} \leq df < M\ r^{k-1},
\Columnsw{\hspace{5mm}}{\]\[}
\left(r = max\left(L, \frac{1}{M}\right)^{1/C}\right).\]

(
\EJ{%
As an exception, $df\ =\ M$ is classified into class 1.
}{%
- $df = M$  Class 1 
})

\Scomment{
\EJ{
If the number $C$ of classes is set to 5 and the lower boundary $L$
is set to 1/32,
and if the maximum document frequency $M$ is larger than or equal to 32
the definition of the frequency classes are as follows:
}{%
- $C\ =\ 5$, $L\ =\ 1/32$ 
- $M\ \geq\ 32$ 
}\\[3\rmm]
\centerline{
\begin{tabular}{@{\EJ{Class}{-}\ }l@{:\hspace{3\rmm}}c@{$\ \leq\ df\ $}c@{$\ $}c}
1&$M$/2&$\leq$&$M$,\\
2&$M$/4&$<$&$M$/2,\\
3&$M$/8&$<$&$M$/4,\\
4&$M$/16&$<$&$M$/8,\\
5&$M$/32&$<$&$M$/16.
\end{tabular}}
\EJ{}{-}
}

\subsection{\EJ{%
Balance Tuning
}{%
-
}}

\EJ{
Another advante of using frequency classes is that
we can change the balance of high-frequency and low-frequency topic words.
The balance parameter $\paramB$, which takes a value within the
interval [-1, 1], is used for this purpose.
The upper limit $N_{\paramB}(k)$ of the number of topic words
taken from frequency classes 1 through $k$ is determined by
\[N_{\paramB}(k) = N \times \left\{\ \paramB \times
\left(\frac{k}{C}\right)^2 +\ (1-\paramB) \times
\left(\frac{k}{C}\right) \right\} .\]
When this balance parameter takes a negative value (such as -1),
that means higher weights are attached to high-frequency words
(or the words in the frequency classes of small class numbers),
and when the parameter takes a positive value (such as 1),
higher weights are attached to the low-frequency words
(or the words in the frequency classes of large class numbers).
}{%
-
-
-
-
-($\paramB$)
- $[-1, 1]$ 
-
-
-
-($\paramB$)
- 1  $k$ 
($N_{\paramB}(k)$) -
\[N_{\paramB}(k) = N \times \left\{\ \paramB \times
\left(\frac{k}{C}\right)^2 +\ (1-\paramB) \times
\left(\frac{k}{C}\right) \right\}\]
}

\EJ{%
In the above definition we determined the number of topic words 
taken from each frequency class indirectly using the accumulated 
number over classes 1 to k.
The reason we did not determine the number directly for individual
classes is that in some cases some frequency classes do not contain
enough words. 
In such cases the above definition allows us to make a compensation
by taking extra topic words from the next frequency class.
}{%
-
-
-
-
}

\Scomment{
\EJ{%
Because the above definition is difficult to understand,
we show in Fig. {\fignumParamB} 
the graphs of standard number of topic words alloted to
each frequency class for various values of balance parameter \paramB.
}{%
-
-
- \fignumParamB\  ($\paramB$) 
1, 0.5, 0, -1 -
}
}

\Scomment{
\Figsw{
\vskip3mm
\begin{figure}
\centerline{\input{\tmpdirA/fig_ParamB.tex}}
\end{figure}
\vskip3mm
}{\centerline{(Fig. \fignumParamB)}}
}

\Scomment{
When the parameter \paramB is positive, higher weights are attached to
high frequency-words;
and when \paramB is negative, higher weights are attached to 
low-frequency words.
When $\paramB = 0$, almost equal numbers of topic words are taken from
each class.
}

\subsection{\EJ{%
Example
}{%
-
}}

\tcomment{\setcounter{section}{5}\setcounter{subsection}{2}
\def\EJ#1#2{#1}\def\fcomment#1{}
\def\tcomment#1{}
\subsection{\EJ{%
Example
}{%
-
}}
}

\EJ{
A primary benefit of using frequency classes in topic word extraction 
is that we can adjust the balance of common topic words and specific 
topic words, and this balance affects the impression made by 
topic word graphs.
}{%
-
-
-
-
}

\EJ{
Figure {\fignumFreqClass} shows six different topic word graphs
generated from a single set of documents retrieved by using \q{ASEAN}.
as the query. The six cases are as follows.
}{%
-\fignumFreqClass  {\em ASEAN} 
-
-
}

\begin{itemize}
\EJ{
\item[] (a) No use of frequency classes.
\item[] (b) Only frequency class 1 is used.
\item[] (c) Frequency classes 1 and 2 are used.
\item[] (d) Frequency classes 1 through 3 are used.
\item[] (d') Same as (d), with balance parameter \paramB = -1.0.
\item[] (d'') Same as (d), with balance parameter \paramB = +1.0.
}{%
\item[] (a) -
\item[] (b) -
\item[] (c) -
\item[] (d) -
\item[] (d') -(d)\paramB=-2
\item[] (d'') -(d)\paramB=+2
}
\end{itemize}

\EJ{%
}{%
-
-
-
-
}

\EJ{%
Let us first examine the topic word graphs for cases (a) and (b),
respectively the case where frequency classes are not used and
the case where only frequency class 1 is used.
These two graphs are very similar, with the small difference that
\q{Nordom} in case (a) is replaced by \q{Rouge} in case (b).
For comparison, the frequency threshold (freq.= 19) between classes 1
and 2 is shown in (a) by a dotted horizontal line.
Since the line is almost at the bottom of the graph,
we know that if we do not use frequency classes most of the topic words
are taken from frequency class 1.
In other words, the case of no use of frequency classes can be well
simulated by the case in which only high frequency classes are used.
}{%
-$\cdots$(a)
-$\cdots$(b)
-
(a)-
-
- (a) 
-(a)(b)
{\em Nordon} - {\em Rouge} 
-(a)
-
}

\EJ{%
Graph (c) is the topic word graph generated when topic words are 
taken from the upper two classes, 1 and 2.
Since the balance parameter {\paramB} is set to 0 (neutral),
almost same numbers of topic words are taken from both classes.
(Seven from class 1 and eight from class 2.)
The dotted horizontal line in the middle of the graph shows the
frequency threshold (freq.=19).
Since the vertical coordinate is proportional to the logarithm of
frequency, the threshould line is located almost at the center.
The words belonging to class 1 are \q{ASEAN} (the query word), and
some related country names such as \q{Thailand} and \q{Singapore},
and this selection is similar to the case in cases (a) and (b).
Some words related to Cambodia, however, such as \q{Cambodia}
and \q{Khmer Rouge}, which belong to class 1 in case (b),
are substituted by other Cambodia-related words, such as
\q{Hun Sen} or \q{Phnom Pehn}, belonging to class 2.
This means that topic words related to a specific topic, in this 
case Cambodia rather than the general topic ASEAN, are more 
likely to appear in case (b) than in case (a).
}{%
-(c)
-$(b=0)$
-
-
-
- {\em ASEAN} 
{\em Singapore} - {\em Thailand} 
(a)-(b)
-
-(b)
{\em Cambodia} - {\em Khmer Rouge} 
{\em Hun Sen} - {\em Phnom Penh} 
}

The graph (d) is topic word graph when the three frequency classes 1 to 
3 are used.
Here again, the balance parameter is set to 0 (neutral),

\Figsw{

\twocolumn[\centerline{\tcomment{\def\fcomment#1{}
\onecolumn
\def\tcomment#1{}}

{\rmm=1mm \unitlength=1\rmm
\def\tmptopic{ASEAN}
\def\tmpframe{
 \multiput(-37.5,0)(75,0){2}{\line(0,1){54}}
 \multiput(-37.5,0)(0,54){2}{\line(1,0){75}}}
\def\tmpthresh#1{\multiput(-37.5,#1)(2,0){38}{\line(1,0){1}}}
\figureformEPSFaa{160}{196}{\fignumFreqClass}{%
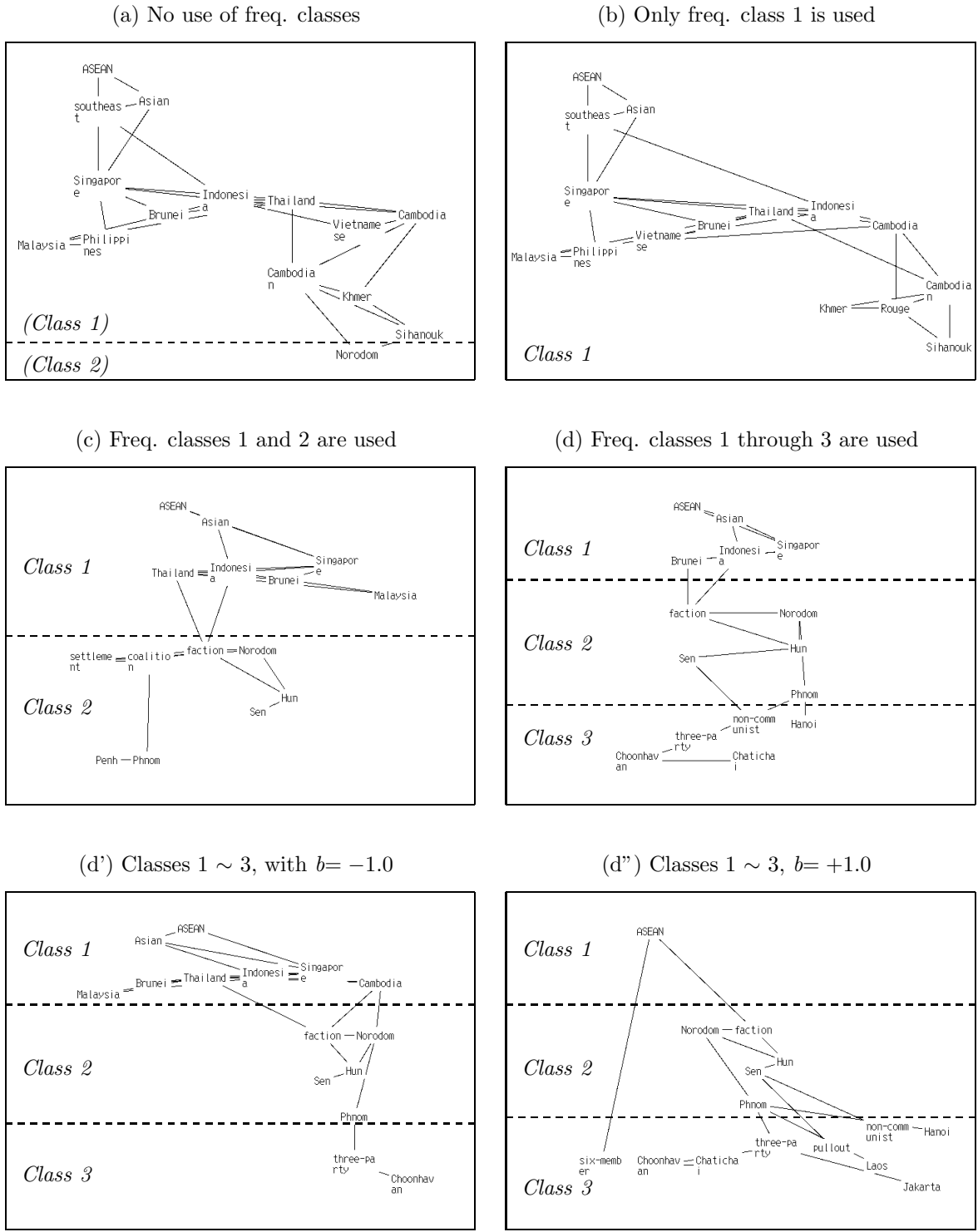
\EJ{Effect of the use of frequency classes}{-}}{%
\put(-40,163){\makebox(0,0)[c]{
  \figureformEPSFba{75}{54}{\tmpdirA/epsf/\tmptopic_b-.epsf}{
    \tmpframe\tmpthresh{6}
    \put(-35,8){\makebox(0,0)[bl]{\it (Class 1)}}
    \put(-35,4){\makebox(0,0)[tl]{\it (Class 2)}}
   }}}
\put( 40,163){\makebox(0,0)[c]{
  \figureformEPSFba{75}{54}{\tmpdirA/epsf/\tmptopic_b+1.epsf}{
    \tmpframe
    \put(-35,3){\makebox(0,0)[bl]{\it Class 1}}
   }}}
\put(-40,95){\makebox(0,0)[c]{
  \figureformEPSFba{75}{54}{\tmpdirA/epsf/\tmptopic_b+2.epsf}{
    \tmpframe\tmpthresh{27}
    \put(-35,37){\makebox(0,0)[bl]{\it Class 1}}
    \put(-35,17){\makebox(0,0)[tl]{\it Class 2}}
   }}}
\put( 40,95){\makebox(0,0)[c]{
  \figureformEPSFba{75}{54}{\tmpdirA/epsf/\tmptopic_b+3.epsf}{
    \tmpframe\tmpthresh{36}\tmpthresh{16}
    \put(-35,40){\makebox(0,0)[bl]{\it Class 1}}
    \put(-35,26){\makebox(0,0)[l]{\it Class 2}}
    \put(-35,12){\makebox(0,0)[tl]{\it Class 3}}
   }}}
\put(-40,27){\makebox(0,0)[c]{
  \figureformEPSFba{75}{54}{\tmpdirA/epsf/\tmptopic_b+3-2b.epsf}{
    \tmpframe\tmpthresh{36}\tmpthresh{17}
    \put(-35,45){\makebox(0,0)[l]{\it Class 1}}
    \put(-35,26){\makebox(0,0)[l]{\it Class 2}}
    \put(-35,9){\makebox(0,0)[l]{\it Class 3}}
   }}}
\put(+40,27){\makebox(0,0)[c]{
  \figureformEPSFba{75}{54}{\tmpdirA/epsf/\tmptopic_b+3+2b.epsf}{
    \tmpframe\tmpthresh{36}\tmpthresh{18}
    \put(-35,45){\makebox(0,0)[l]{\it Class 1}}
    \put(-35,26){\makebox(0,0)[l]{\it Class 2}}
    \put(-35,8){\makebox(0,0)[tl]{\it Class 3}}
   }}}
\put(-40,196){\makebox(0,0)[t]{\EJ{(a)\  No use of freq. classes}{}}}
\put( 40,196){\makebox(0,0)[t]{\EJ{(b)\  Only freq. class 1 is used}{}}}
\put(-40,128){\makebox(0,0)[t]{\EJ{(c)\  Freq. classes 1 and 2 are used}{}}}
\put( 40,128){\makebox(0,0)[t]{\EJ{(d)\  Freq. classes 1 through 3 are used}{}}}
\put(-40, 60){\makebox(0,0)[t]{\EJ{(d')\  Classes 1 $\sim$ 3, with \paramB $= -1.0$}{}}}
\put( 40, 60){\makebox(0,0)[t]{\EJ{(d'')\  Classes 1 $\sim$ 3, \paramB $= +1.0$}{}}}
}}
}\vskip5mm]

}{
\centerline{(Fig. \fignumFreqClass)}
}

\EJ{%
and each five words are taken from three classes.
This time, five new words in class 3 -- \q{Hanoi},
\q{non-communist}, \q{three-party}, and \q{Chatichai Choonhavan}
(the then prime minister of Thailand) --
are added in place of \q{Thailand} and \q{Malaysia} in class 1
and \q{settlement}, \q{coalition}, and \q{Penh} in class 2.
This change is interpreted as showing that we get still-more-specific
words related to a major specific topic rather than a general topic.
}{%
-(d)
-$(b=0)$
-
(c)- {\em Thailand}  {\em Malaysia}
- {\em settlement}  {\em coallition}, {\em Penh} 
- {\em Hanoi}, {\em non-communist},
{\em three-party}, {\em Chatichai Choonhavan}
(the then prime minister of Thailand) -
}

\EJ{%
\newpage
The cases (d') and (d'') are same as case (d) with respect to
the frequency classes but with different balance parameters.
In case (d') the balance parameter {\paramB} is set to -1.0 and
more weight is on the common words in class 1.
Actually, the numbers of topic words taken from classes 1 to 3 are
8, 5, and 2 respectively.
Since the weight of class 3 is very small,
the graph is similar to the case (c) (or even (a) or (b)).
Conversely, in case (d'') the balance parameter is set to 1.0 and
more weight is on the specific words in class 3.
This time, nine words are taken from class 3, 5 words from class 2
and only 1 word from class 1.
Since most of words are taken from class 3,
the topic word graph seems to outline a major specific problem
rather than whole articles concerning \q{ASEAN}.
(In this case the major specific problem is the peace talks
at Jakarta by Cambodia's Hun Sen government and three guerrilla groups.)
}{%
-(d')(d'')(d)
-
-(d')(\paramB)$-2$
-
-
-(a)(b)
-
(d'')-(\paramB)$+2$
-
-
-ASEAN 
-
-
-(Khmer Rouge )
-
- ASEAN 
-
}


\tcomment{\setcounter{section}{6}\setcounter{page}{7}\def\tcomment#1{}}

\section{Conclusion} 

\def\tmp{\addtocounter{enumi}{1}{$\cdot$}}
\setcounter{enumi}{0}

To make an interactive guidance mechanism for document 
retrieval systems, we developed a user-interface which presents users 
a visualized graph of topic words at each stage of the retrieval process.
Topic words are automatically extracted by frequency analysis and 
the relationship between topic words is measured by their co-occurrence.

We built a prototype retrieval system for about 80 thousand articles of
AP Newswire '89, and
experiments with this system support our expectation that the guidance 
provided by the topic word graph is useful for interactive screening.
By using the topic word extraction method using frequency classes,
we succeeded in taking well-balanced topic words from the wide range 
of word frequencies.
This method is also advantageous for adjusting the balance of the
high-frequency topic words and low-frequency topic words, a balance
that greatly affects the user's impression of the topic word graph.

\Comment{
\subsection*{Future Work}
\setcounter{enumi}{0}

Proper nouns and compound nouns are important as keywords, but they
are mostly unregistered in dictionaries.
Currently, most of these words are treated as a sequence of regular words.
Automatic recognition and collection of such words are needed for further
progress.

As for the criterion for linking topic words, there are many possibilities.
In the current system, each word X is linked to word Y, which has a
higher document frequency than X and has the highest 
co-occurrence with X ($\rm F_{xy}$/$\rm F_y$ See Sect. 2.2.).
We need to compare this criterion with others and search for the best one.
}
\def\Fbib#1{#1}

\Fdouble{\singlespace\rnormalsize}

\Fbib{%
\Figsw{}{\vfil\break}
\bibliographystyle{\bibdir/fullname}
\bibliography{\bibdir/niwa}
}

\Figsw{}{\newpage\input{figs}}

\end{document}